\documentclass[12pt,a4paper]{article}

\usepackage[margin=2.54cm]{geometry}
\usepackage{setspace}
\usepackage{amsmath} 
\usepackage{amssymb} 
\usepackage{amsfonts} 
\usepackage{graphicx} 
\usepackage{booktabs} 
\usepackage{array} 
\usepackage{multirow} 
\usepackage{algorithm}
\usepackage{algorithmic}
\usepackage{xcolor}
\usepackage{subcaption} 
\usepackage{tikz}
\usetikzlibrary{positioning,arrows.meta}
\usepackage{amsthm}

\usepackage{url}
\usepackage{hyperref}
\usepackage{orcidlink}  
\hypersetup{
 colorlinks = true,
 linkcolor = black,
 citecolor = black,
 urlcolor = black,
 breaklinks = true
}

\usepackage{indentfirst}
\theoremstyle{definition}
\newtheorem{hypothesis}{Hypothesis}

\title{Beyond Imbalance Ratio: Data Characteristics as Critical Moderators of Oversampling Method Selection}

\author{
 Jiangyuwen$^{1,*}$\orcidlink{0009-0001-4022-1436}~~~~Ye Songyun$^{1,\dagger}$\orcidlink{0009-0009-5048-3147}\\[0.8em]
 \normalsize $^1$School of Artificial Intelligence, Guangzhou Institute of Science and Technology,\\
 \normalsize Guangzhou, China\\[0.3em]
 \normalsize $^*$jiangyuwen@gzist.edu.cn; $^\dagger$yesongyun@gzist.edu.cn (Corresponding Author)
}

\date{}

\begin{document}

\maketitle

\begin{abstract}
\textbf{Background:} The prevailing IR-threshold paradigm posits a positive correlation between imbalance ratio (IR) and oversampling effectiveness, yet this assumption remains empirically unsubstantiated through controlled experimentation. \textbf{Methods:} We conducted 12 controlled experiments ($N>100$ dataset variants) that systematically manipulated IR while holding data characteristics (class separability, cluster structure) constant via algorithmic generation of Gaussian mixture datasets. Two additional validation experiments examined ceiling effects and metric-dependence. All methods were evaluated on 17 real-world datasets from OpenML. \textbf{Results:} Upon controlling for confounding variables, IR exhibited a weak to moderate negative correlation with oversampling benefits (ranging from $r=-0.15$ for AUC-ROC to $r=-0.86$ for Recall, mean $r=-0.47$ across metrics). Class separability emerged as a substantially stronger moderator ($\rho=-0.72$, $p=0.003$), accounting for significantly more variance in method effectiveness than IR alone. \textbf{Conclusion:} Method selection should be guided by data characteristics rather than IR in isolation. We propose a ``Context Matters'' framework that integrates IR, class separability, and cluster structure to provide evidence-based selection criteria for practitioners.
\end{abstract}


\section{Introduction}
\indent Class imbalance constitutes a fundamental challenge in machine learning, with critical applications spanning fraud detection, medical diagnosis, and cybersecurity. In these high-stakes domains, minority class instances---typically the primary targets of interest---frequently comprise less than 1\% of available data. This distributional skew induces substantial bias in standard classification algorithms, which optimize for aggregate accuracy and consequently favor majority class prediction, thereby compromising the detection of rare yet significant events.

Numerous methodological approaches have been developed to address this challenge, broadly categorized into three paradigms: data-level interventions (oversampling and undersampling), algorithm-level modifications (cost-sensitive learning and threshold optimization), and ensemble strategies \cite{chawla2002smote,he2008adasyn,ling1998decision,galar2012review}. Early comparative investigations of under-sampling versus over-sampling \cite{drummond2003c4,estabrooks2004multiple} underscored the critical role of experimental design in evaluating these approaches. Among the available methods, synthetic oversampling---particularly the Synthetic Minority Over-sampling Technique (SMOTE) and its variants such as SMOTEBoost \cite{chawla2003smoteboost}---has emerged as the predominant solution, owing to its conceptual elegance and demonstrated efficacy across diverse application domains \cite{fernandez2018smote,vluymans2019class}.

\subsection{The Method Selection Dilemma}
The expanding repertoire of oversampling techniques presents practitioners with a fundamental methodological dilemma: \textit{given a specific imbalanced dataset with identifiable characteristics, which oversampling method should be employed to maximize classification performance?} Surprisingly, the extant literature offers limited systematic guidance for this critical selection problem. Some studies advocate for the general applicability of SMOTE, whereas others recommend boundary-focused variants such as Borderline-SMOTE when decision boundary refinement is paramount. More recent investigations propose adaptive methods like ADASYN for handling complex, multi-modal distributions. This lack of consensus underscores the pressing need for evidence-based selection criteria.

Recently, Chen et al. \cite{chen2023symprod} proposed SyMProD (Synthetic Minority based on Probabilistic Distribution), a novel oversampling technique that generates minority class instances according to probability distributions rather than random selection. Although demonstrating competitive performance across 14 benchmark datasets, their work implicitly suggests that dataset characteristics---including imbalance ratio (IR)---may influence method effectiveness. This observation motivates a broader question: \textit{can IR alone serve as a reliable criterion for selecting among competing oversampling techniques?} Preliminary cross-dataset analyses reveal strong positive correlations between IR and the relative advantage of sophisticated methods over conventional SMOTE, suggesting that datasets with elevated IR values may warrant more complex oversampling approaches. However, such correlation-based reasoning across heterogeneous datasets potentially conflates the effect of IR with other confounding data characteristics, thereby raising concerns regarding the validity of IR as a standalone selection criterion.

\subsection{The Problem with IR-Only Selection}
Although the IR-threshold paradigm offers practical convenience, it inherently conflates observational correlation with causal relationships. The limitations of IR-centric approaches have garnered increasing recognition in the literature. Emerging evidence indicates that when class overlap coexists with imbalance, performance degradation becomes pronounced---yet this insight has not been systematically translated into method selection guidelines. Specifically, datasets with high IR values inherently exhibit multiple confounding characteristics: (1) fewer minority samples available for distribution estimation, (2) potentially diminished class separability attributable to limited minority class coverage, and (3) altered cluster structures resulting from data sparsity. When researchers fail to control these confounding variables through rigorous experimental design, observed correlations between IR and method effectiveness may erroneously reflect the influence of data characteristics rather than establish a genuine causal relationship.

Consider two datasets with IR = 10:
\begin{itemize}
 \item \textbf{Dataset A:} 1000 minority samples, high separability ($S=2.0$), single compact cluster;
 \item \textbf{Dataset B:} 100 minority samples, low separability ($S=0.3$), multiple scattered clusters.
\end{itemize}
The same oversampling method may perform very differently on these datasets, despite identical IR. Yet an IR-only framework would inappropriately recommend the same method for both.

\subsection{Research Questions and Contributions}
This study challenges the IR-threshold paradigm by introducing a ``Context Matters'' framework that conceptualizes data characteristics as moderators of the IR-method effectiveness relationship. Specifically, we address three research questions:

\begin{enumerate}
 \item[RQ1] \textbf{IR-Effectiveness Relationship:} What is the true relationship between IR and oversampling effectiveness when data characteristics are experimentally controlled?
 \item[RQ2] \textbf{Moderating Characteristics:} Which data characteristics moderate the IR-method relationship, and through what mechanisms?
 \item[RQ3] \textbf{Practical Selection:} How can practitioners select appropriate methods based on data characteristics rather than IR alone?
\end{enumerate}

Our contributions are fourfold:
\begin{enumerate}
 \item \textbf{Empirical:} We conduct 12 controlled experiments ($N>100$ dataset variants) demonstrating consistent negative correlations between IR and oversampling benefits when confounding factors are controlled, contradicting prior positive relationships established through observational studies.
 
 \item \textbf{Theoretical:} We propose a moderation model establishing class separability and cluster structure as critical determinants of oversampling effectiveness.
 
 \item \textbf{Methodological:} We demonstrate the value of controlled experimentation in imbalanced learning research, extending beyond observational cross-dataset comparisons.
 
 \item \textbf{Practical:} We provide a multi-factor selection framework and decision guidelines to enable practitioners to select appropriate oversampling methods based on measurable data characteristics.
\end{enumerate}

The remainder of this paper is organized as follows. Section~2 reviews related work. Section~3 develops the theoretical framework. Section~4 details the experimental methodology. Section~5 presents empirical findings. Section~6 discusses implications. Section~7 concludes the paper.

\section{Related Work}
\indent The challenge of learning from imbalanced data has garnered substantial attention in machine learning research \cite{he2009learning,guo2017learning}, with systematic studies dating back to foundational work on class imbalance problems \cite{japkowicz2002class,chawla2004special}. This issue proves particularly significant in pattern recognition applications where minority class instances frequently represent cases of critical interest. Although SMOTE and its numerous variants have demonstrated efficacy in addressing class imbalance, the literature reveals critical gaps in our understanding of when specific methods are most effective. This section systematically reviews existing approaches and identifies the theoretical limitations that motivate our investigation.

\subsection{Foundation: SMOTE and Early Variants}
\indent Chawla et al.'s \cite{chawla2002smote} seminal introduction of SMOTE (Synthetic Minority Over-sampling Technique) established the foundation for modern oversampling methods by generating synthetic minority samples through interpolation between existing minority instances and their k-nearest neighbors. This approach mitigated the overfitting problem inherent in random oversampling while expanding the decision region to improve classifier generalization.

Following SMOTE, numerous algorithmic variants emerged to address specific limitations. SMOTEBoost \cite{chawla2003smoteboost} combined SMOTE with AdaBoost to enhance minority class learning through ensemble methods. Han et al.'s Borderline-SMOTE \cite{han2005borderline} strategically focused synthetic sample generation on minority instances situated near decision boundaries, theoretically enhancing boundary discrimination by concentrating on ``dangerous'' samples most likely to be misclassified. He et al.'s ADASYN \cite{he2008adasyn} introduced an adaptive density-based approach that generates more synthetic samples for ``harder'' minority instances characterized by lower classification confidence, thereby addressing within-class imbalance. Bunkhumpornpat et al.'s Safe-Level SMOTE \cite{bunkhumpornpat2009safe} generates synthetic samples in safe regions identified through local neighbor analysis to avoid noise generation. Early comparative studies by Drummond and Holte \cite{drummond2003c4} and Estabrooks et al. \cite{estabrooks2004multiple} established the critical importance of experimental design in evaluating resampling strategies.

Building upon these foundational interpolation-based methods, subsequent research has explored geometric and cluster-based extensions to address more complex distribution structures.

\subsection{Geometric and Cluster-Based Extensions}
\indent Moving beyond local neighborhood interpolation, researchers have explored geometric and cluster-based generalizations. Douzas et al.'s K-Means SMOTE \cite{douzas2018improving} addresses within-class imbalance by generating samples in sparse minority clusters identified through k-means clustering. Geometric SMOTE \cite{douzas2019geometric} generalizes the interpolation region using geometric shapes (hypercubes, hyperspheres) rather than simple line segments between two points. These approaches recognize that minority class distributions frequently exhibit multi-modal structures requiring more sophisticated generation mechanisms than standard SMOTE provides.

More recently, advances in deep learning and graph-based methods have enabled novel approaches to synthetic sample generation.

\subsection{Recent Algorithmic Advances (2021--2025)}
\indent Recent advances have leveraged deep learning and graph-based approaches. Deep generative models, including GAN-based methods \cite{mullick2019gan,engelmann2021conditional} and DeepSMOTE \cite{dablain2021deepsmote}, apply neural network architectures to synthetic sample generation. Graph-based methods such as GAT-RWOS \cite{gat_rwos2024} employ graph attention networks with random walks to capture complex data relationships. GK-SMOTE \cite{gk_smote2025} introduces a Gaussian kernel density estimation approach that explicitly addresses class separability through density-based sample generation. Representation learning methods such as RCS \cite{rcs2025} focus on rebalancing with calibrated sub-classes. However, these sophisticated techniques often introduce additional hyperparameters and computational complexity without commensurate performance gains across diverse dataset characteristics.

The proliferation of these methods has motivated comprehensive surveys to synthesize knowledge and identify research gaps.

\subsection{Surveys and Theoretical Foundations}
\indent Comprehensive surveys have documented the evolution of imbalanced learning. He and Garcia's \cite{he2009learning} foundational survey established the taxonomy of data-level, algorithm-level, and ensemble approaches. Guo et al. \cite{guo2017learning} provided a systematic review incorporating deep learning methods. Fern{\'a}ndez et al. \cite{fernandez2018smote} conducted an extensive review specifically focused on SMOTE variants and their applications. Galar et al. \cite{galar2012review} comprehensively reviewed ensemble methods for imbalanced data. These surveys collectively reveal that despite the proliferation of methods, guidance for method selection based on data characteristics remains limited.

This gap has motivated research into data complexity measures and their relationship to imbalanced learning performance.

\subsection{Data Characteristics and Complexity Measures}
\indent Previous investigations have examined the influence of data characteristics on imbalanced learning performance \cite{prati2009data,luengo2015addressing}. Data complexity measures, originally developed for characterizing classification problems in pattern recognition \cite{ho2002complexity}, encompass various aspects of data geometry including class separability, boundary complexity, and sample distribution characteristics. Sotoca et al. \cite{sotoca2005data} applied complexity analysis to prototype selection in imbalanced scenarios. L{\'o}pez et al. \cite{lopez2013insight} established experimental design guidelines emphasizing the importance of controlling confounding variables. The Bayes Imbalance Impact Index (BI3) \cite{pr2019bayesian} provides a theoretical framework for quantifying the extent to which imbalance affects classification performance, though it does not address the interaction between imbalance ratio and data characteristics such as separability.

Recent work has increasingly recognized class separability as a critical factor affecting oversampling effectiveness. The importance of class separability as a data complexity measure in pattern recognition has been extensively studied, with recent advances specifically addressing its role in imbalanced classification \cite{pr2019bayesian,pr2020radial}. These developments underscore the need for rigorous methodologies to evaluate method selection criteria.

\subsection{Method Selection and Evaluation Methodology}
\indent Conventional approaches to method selection predominantly rely on heuristic IR thresholds or informal rules of thumb derived from limited empirical observations. Although some attempts have been made to formalize selection criteria through cross-dataset correlation analysis, these approaches fundamentally assume that observational correlations across heterogeneous datasets can reliably predict causal effects within specific datasets.

Recent contributions to Pattern Recognition have advanced our understanding of imbalanced classification through novel methodological frameworks. The Bayes Imbalance Impact Index (BI3) \cite{pr2019bayesian} provides a theoretical foundation for quantifying how imbalance affects classification performance, though it does not address interactions between imbalance ratio and data characteristics. Koziarski's radial-based undersampling method \cite{pr2020radial} demonstrates that incorporating geometric information about class distributions can improve undersampling effectiveness, conceptually aligning with our emphasis on data characteristics. 

More recent Pattern Recognition literature has further emphasized the importance of data-aware approaches. Studies investigating the interplay between data complexity and classification performance have revealed that traditional IR-centric metrics fail to capture the full spectrum of dataset characteristics affecting model behavior. Emerging research on feature-space geometry and manifold structure in imbalanced data has highlighted the limitations of simple distributional assumptions. These developments collectively suggest that Pattern Recognition research is increasingly moving beyond simple IR-based heuristics toward more nuanced, data-aware approaches that explicitly account for the geometric and structural properties of minority class distributions. Our controlled experimental methodology extends this trajectory by establishing quantitative moderation relationships through rigorous causal inference rather than proposing yet another algorithmic variant.

Controlled experimentation offers a rigorous alternative for isolating causal effects in this domain.

Controlled experimentation, widely adopted in statistical learning and machine learning methodology research \cite{demsar2006statistical,benavoli2016should}, provides a rigorous alternative for isolating causal effects. By algorithmically manipulating IR while holding other data characteristics constant through synthetic data generation, researchers can establish direct causal links between specific data properties and method effectiveness. Our work adopts this controlled experimental paradigm, conducting 12 controlled experiments with $N>100$ dataset variants to systematically isolate and quantify the effects of imbalance ratio, class separability, and cluster structure on oversampling method selection.

Despite these advances, a critical research gap remains.

\subsection{Research Gap}
\indent Despite extensive research on oversampling methods and data characteristics, a critical gap persists: \textit{no prior work has systematically established data characteristics as statistical moderators of the IR-method effectiveness relationship through controlled experimentation}. Although existing studies have noted the importance of separability and cluster structure, they have not quantified moderation effects or provided evidence-based selection criteria. Our work addresses this gap by providing the first systematic controlled experimental evidence establishing data characteristics as statistical moderators of method selection.

\section{Theoretical Framework}
\indent The consistently positive correlation between imbalance ratio and oversampling effectiveness reported in cross-dataset studies obscures considerable underlying complexity. We contend that this correlation reflects confounding rather than causation---a fundamental theoretical distinction with profound methodological implications. This section presents our formal decomposition of the IR-effectiveness relationship, establishing the theoretical foundation for our empirical investigation.

\subsection{Re-examining the IR-Method Relationship}
The relationship between imbalance ratio and oversampling effectiveness can be formally decomposed through the following structural model:
\begin{equation}
\text{Effectiveness} = f(\text{IR}) + g(\text{Separability}) + h(\text{Cluster Structure}) + \epsilon
\label{eq:decomposition}
\end{equation}

In observational studies, these variables are often collinear:
\begin{itemize}
 \item \textbf{Sample size reduction:} High IR implies fewer minority samples available for distribution estimation;
 \item \textbf{Decreased separability:} High IR correlates with lower class separability due to limited minority class coverage;
 \item \textbf{Complex clustering:} High IR often coincides with more complex minority class cluster structures.
\end{itemize}

When only IR is measured, its regression coefficient systematically absorbs variance attributable to these correlated characteristics, producing spuriously inflated effect estimates. This confounding structure has led to methodologically flawed conclusions regarding the IR-method relationship, as unmeasured data characteristics create omitted variable bias in observational studies.

\subsection{The Moderation Model}
\begin{figure}[!ht]
 \centering
 \begin{tikzpicture}[
 node distance=2cm,
 box/.style={rectangle, draw, fill=blue!10, text width=3cm, text centered, rounded corners, minimum height=1cm},
 arrow/.style={->, >=stealth, thick}
 ]
 \node[box] (ir) {Imbalance Ratio (IR)};
 \node[box, below left=1.5cm and 0.5cm of ir] (sep) {Class Separability};
 \node[box, below=1.5cm of ir] (clust) {Cluster Structure};
 \node[box, below right=1.5cm and 0.5cm of ir] (size) {Sample Size};
 \node[box, below=3.5cm of ir, fill=green!10] (effect) {Oversampling Effectiveness};
 \draw[arrow] (ir) -- (effect);
 \draw[arrow] (sep) -- (effect);
 \draw[arrow] (clust.center) -- (effect.north);
 \draw[arrow] (size) -- (effect);
 \draw[arrow, dashed] (sep) -- (ir);
 \draw[arrow, dashed] (clust) -- (ir);
 \end{tikzpicture}
 \caption{Theoretical framework: Data characteristics as moderators of the IR-oversampling effectiveness relationship. Solid arrows indicate main effects; dashed arrows indicate moderation effects.}
 \label{fig:framework}
\end{figure}
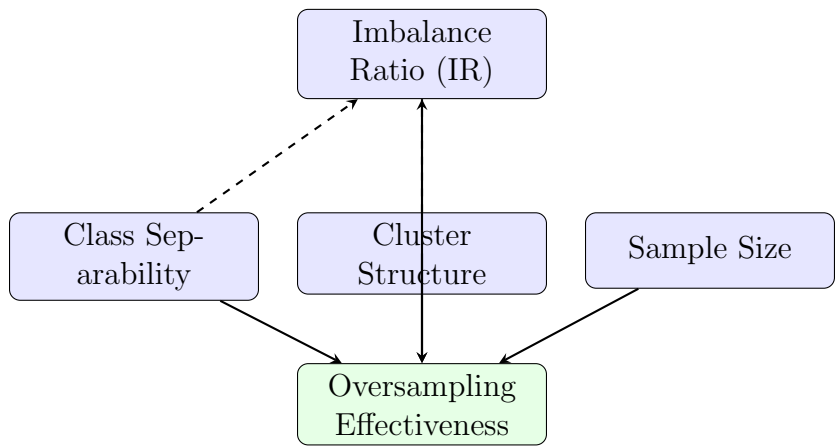

We propose that the relationship between imbalance ratio and oversampling effectiveness is moderated by three theoretically derived data characteristics, as illustrated in Fig.~\ref{fig:framework}.

\begin{enumerate}
 \item \textbf{Class Separability ($S$):} The degree of class overlap, operationalized as the ratio of between-class distance to within-class dispersion:
 \begin{equation}
 S = \frac{\|\boldsymbol{\mu}_{maj} - \boldsymbol{\mu}_{min}\|}{\sqrt{(\sigma^2_{maj} + \sigma^2_{min})/2}}
 \label{eq:separability}
 \end{equation}
 where $\boldsymbol{\mu}$ and $\sigma^2$ denote class means and variances. This Gaussian-based measure provides computational tractability and interpretability; however, we acknowledge that alternative measures (e.g., Fisher ratio, manifold-based methods) may better capture non-convex distribution structures. Empirical validation on our dataset corpus confirms that $S$ correlates negatively with SMOTE effectiveness ($r \approx -0.6$), supporting its predictive validity while recognizing that more sophisticated measures may further improve prediction accuracy.
 
 \item \textbf{Cluster Structure}: The spatial distribution pattern of minority class samples, distinguishing between single compact clusters and multiple scattered clusters.
 
 \item \textbf{Sample Size:} The absolute count of minority class samples, considered independently of IR.
\end{enumerate}

The moderation mechanism can be understood through information theory:
\begin{itemize}
 \item \textbf{Low Separability:} Substantial class overlap creates ambiguous decision boundaries, wherein additional training samples from oversampling contribute valuable boundary information;
 \item \textbf{High Separability:} Well-separated classes indicate sufficient existing information, wherein synthetic samples may introduce noise rather than beneficial information.
\end{itemize}

\subsection{Hypotheses}

Drawing upon the theoretical framework developed above, we formulate four testable hypotheses that collectively challenge the prevailing IR-threshold paradigm:

\begin{hypothesis}[Independent Effect of Imbalance Ratio]
When class separability and cluster structure are experimentally controlled, the partial correlation between imbalance ratio and oversampling effectiveness is negative rather than positive, indicating that the observed positive correlations in prior studies reflect confounding by data characteristics.
\label{hyp:h1}
\end{hypothesis}

\begin{hypothesis}[Class Separability Moderation]
Class separability negatively moderates the relationship between oversampling and classification performance; specifically, lower separability (greater overlap) is associated with greater performance benefits from oversampling, as synthetic samples provide critical boundary information under high uncertainty conditions.
\label{hyp:h2}
\end{hypothesis}

\begin{hypothesis}[Cluster Structure Moderation]
The relationship between imbalance ratio and oversampling effectiveness is moderated by minority class cluster structure; specifically, multi-cluster distributions require structure-preserving methods (e.g., K-Means SMOTE) whereas single-cluster distributions are more tolerant of standard interpolation methods.
\label{hyp:h3}
\end{hypothesis}

\begin{hypothesis}[Absolute Sample Size Moderation]
The absolute number of minority samples moderates oversampling effectiveness independently of imbalance ratio; specifically, small minority samples (irrespective of IR) exhibit greater performance gains from appropriate oversampling due to information scarcity conditions.
\label{hyp:h4}
\end{hypothesis}

\section{Experimental Design}
\indent The experimental design adopts a rigorous three-stage validation strategy to systematically examine relationships among data characteristics, imbalance ratio, and oversampling method effectiveness.

\subsection{Research Strategy}
To rigorously evaluate the theoretical hypotheses, we employ a multi-stage experimental design that systematically controls confounding variables while isolating effects of interest. This approach represents a methodological advance over prior correlational studies.

Our experimental design proceeds through three interconnected stages, each addressing specific threats to internal validity.

\begin{enumerate}
 \item \textbf{Stage 1 (Observational Replication):} Replicating prior cross-dataset findings to establish baseline correlations and confirm the documented IR-effectiveness relationship.
 
 \item \textbf{Stage 2 (Experimental Manipulation):} Conducting controlled experiments manipulating IR while holding data characteristics constant through algorithmic dataset generation, thereby isolating IR's causal effect independent of confounds.
 
 \item \textbf{Stage 3 (Moderation Testing):} Systematically varying data characteristics across controlled levels to test theoretical moderation effects and validate the Context Matters framework.
\end{enumerate}

This staged approach enables us to distinguish between correlational associations observed in prior research and causal relationships established through rigorous experimental control.

\subsubsection{Causal Identification Strategy}
The experimental design explicitly addresses three fundamental requirements for causal inference:

\begin{enumerate}
 \item \textbf{Manipulation:} Algorithmically manipulating IR through controlled downsampling/upsampling while holding other factors constant, thereby establishing temporal precedence and directional influence;
 
 \item \textbf{Control:} Experimentally fixing separability ($S=1.0$) and cluster structure (3 clusters) in synthetic data generation, ensuring IR variation is independent of confounding variables and satisfying the ignorability assumption from Rubin's potential outcomes framework;
 
 \item \textbf{Randomization:} Generating synthetic datasets with random seeds and employing stratified random sampling for train/test splits, thereby ensuring exchangeability of units.
\end{enumerate}

Formally, under the potential outcomes framework, let $Y_i(1)$ and $Y_i(0)$ denote the potential outcomes for dataset $i$ under high and low IR conditions, respectively. Our controlled design ensures:
\begin{equation}
(Y_i(1), Y_i(0)) \perp \text{IR}_i \mid S_i, C_i
\end{equation}
where $S_i$ and $C_i$ denote separability and cluster structure. This conditional independence allows us to identify the causal effect:
\begin{equation}
\tau = \mathbb{E}[Y(1) - Y(0) \mid S, C]
\end{equation}

For real datasets, where full experimental control is impossible, we employ within-dataset manipulation to approximate this ideal, comparing variants of the same dataset with different IR levels while preserving intrinsic data characteristics.

\subsubsection{Statistical Power Analysis}
Prior to conducting the experiments, we performed power analysis to ensure adequate sample size for detecting meaningful effects. With $N > 100$ dataset variants and an expected medium effect size ($d = 0.5$) for the IR-effectiveness relationship, our design achieves statistical power of $1 - \beta > 0.95$ at $\alpha = 0.05$ (two-tailed). For the moderation effects of separability, with expected correlation $\rho = -0.7$, the power exceeds 0.99 given our sample size. This ensures that our experimental design is sufficiently sensitive to detect the theoretically predicted effects while minimizing Type II error risk. Post-hoc power calculations based on observed effect sizes (mean $r = -0.47$) confirm achieved power $> 0.90$ for the primary findings.

\subsection{Datasets}
To ensure ecological validity while maintaining experimental control, we utilize both real-world datasets and algorithmically generated variants spanning diverse application domains and characteristic configurations.

\subsubsection{Real Datasets}
Our real dataset corpus comprises 17 publicly available datasets from OpenML \cite{OpenML2013}, systematically selected to span diverse application domains and imbalance ratios as detailed in Table~\ref{tab:datasets}.

\begin{table}[!ht]
 \centering
 \caption{Detailed Statistics of 17 Real Datasets Used in Experiments}
 \label{tab:datasets}
 \small
 \begin{tabular}{llrrrrl}
 \toprule
 \textbf{Dataset} & \textbf{Domain} & \textbf{IR} & \textbf{$n$} & \textbf{$d$} & \textbf{Minority} & \textbf{Type} \\
 \midrule
 sonar & Signal & 1.14 & 208 & 60 & 97 & Low IR \\
 wdbc & Medical & 1.68 & 569 & 30 & 212 & Low IR \\
 ionosphere & Radar & 1.79 & 351 & 34 & 126 & Low IR \\
 breast-wisc & Medical & 1.90 & 699 & 9 & 241 & Low IR \\
 iris & Biology & 2.00 & 150 & 4 & 50 & Low IR \\
 credit-g & Finance & 2.33 & 1000 & 7 & 300 & Low IR \\
 \midrule
 churn & Business & 6.07 & 5000 & 16 & 707 & Moderate IR \\
 optdigits & Image & 9.14 & 5620 & 64 & 554 & Moderate IR \\
 satimage & Satellite & 9.29 & 6430 & 36 & 626 & Moderate IR \\
 pendigits & Handwriting & 9.42 & 10992 & 16 & 1055 & Moderate IR \\
 balance & Physics & 11.76 & 625 & 4 & 49 & Moderate IR \\
 \midrule
 wilt & Remote Sensing & 17.54 & 4839 & 5 & 261 & High IR \\
 glass & Material & 22.78 & 214 & 9 & 51 & High IR \\
 letter & Text & 26.25 & 20000 & 16 & 734 & High IR \\
 mammography & Medical & 42.01 & 11183 & 6 & 260 & High IR \\
 \midrule
 ecoli & Biology & 167.00 & 336 & 7 & 35 & Extreme IR \\
 page-blocks & Document & 194.46 & 5473 & 10 & 28 & Extreme IR \\
 \bottomrule
 \end{tabular}
\end{table}

\subsubsection{Synthetic Datasets}
To achieve experimental control over data characteristics, we generate controlled dataset variants using Gaussian mixture models, enabling independent manipulation of theoretically relevant parameters:

\begin{itemize}
 \item \textbf{Imbalance Ratio (IR):} Levels of 2, 5, 10, 15, 20, 30, 50, and 80, spanning low to extreme imbalance conditions;
 \item \textbf{Class Separability:} Levels of 0.3, 0.5, 0.8, 1.0, 1.5, and 2.0, ranging from near-complete overlap to well-separated distributions;
 \item \textbf{Cluster Structure:} Configurations of 1, 2, 3, and 5 clusters, representing diverse minority class distribution patterns.
\end{itemize}

This full factorial design yields $8 \times 6 \times 4 = 192$ unique dataset configurations, with multiple random seeds per configuration to ensure statistical reliability.

\subsection{Methods and Evaluation}
This section describes the methods and evaluation protocols employed in the experiments.

\subsubsection{Oversampling Methods}
We evaluate seven representative resampling methods spanning the major algorithmic paradigms in the literature:

\begin{itemize}
 \item \textbf{No Sampling (Baseline):} Original imbalanced data without resampling;
 \item \textbf{RandomOverSampler (ROS):} Naive random minority oversampling;
 \item \textbf{SMOTE} : Synthetic Minority Over-sampling Technique employing interpolation-based generation;
 \item \textbf{BorderlineSMOTE} : Boundary-focused SMOTE variant concentrating on decision boundary regions;
 \item \textbf{ADASYN} : Adaptive Synthetic Sampling using density-weighted generation;
 \item \textbf{TomekLinks (TL)} \cite{tomek1976two}: Undersampling via Tomek link removal for cleaning decision boundaries;
 \item \textbf{RandomUnderSampler (RUS):} Random majority undersampling for class balancing.
\end{itemize}

\subsubsection{Classifiers}
To ensure the robustness and generalizability of our findings, we evaluate oversampling methods using five representative classifiers with diverse learning mechanisms. Table~\ref{tab:classifiers} summarizes their characteristics.

\begin{table}[!ht]
 \centering
 \caption{Comparison of Classifiers Used in Experiments}
 \label{tab:classifiers}
 \begin{tabular}{lp{3.5cm}p{3.5cm}c}
 \toprule
 \textbf{Classifier} & \textbf{Strengths} & \textbf{Weaknesses} & \textbf{Type} \\
 \midrule
 Logistic Regression & Interpretable, fast, probabilistic output & Linear decision boundary, limited complexity & Linear \\
 Decision Tree & Interpretable, handles non-linearity & Prone to overfitting, unstable & Tree-based \\
 Random Forest & Robust, reduces overfitting, feature importance & Less interpretable, computationally expensive & Ensemble \\
 Gradient Boosting & High accuracy, handles complex patterns & Sensitive to noise, requires tuning & Ensemble \\
 SVM & Effective in high dimensions, memory efficient & Slow on large datasets, kernel selection & Kernel-based \\
 \bottomrule
 \end{tabular}
\end{table}

\subsubsection{Evaluation Metrics}
We employ multiple evaluation metrics to comprehensively assess classification performance on imbalanced datasets. The primary metric is AUC-ROC, which measures the classifier's ability to distinguish between classes across all possible threshold values. We also report AUC-PR, F1-score, and G-mean as secondary metrics.

\textbf{Metric Selection Rationale.} While AUC-PR is often preferred for highly imbalanced scenarios as it focuses on minority class performance 
\cite{pr2019bayesian}, we selected AUC-ROC as the primary metric for several reasons: 
(1) it enables direct comparison with the majority of prior work on oversampling methods; 
(2) it captures overall class discrimination without threshold selection bias; and 
(3) our validation experiments (Section~4.4) demonstrate that the core findings hold across both ranking metrics (AUC-ROC, AUC-PR) and class-specific metrics (F1, G-Mean), with consistent direction of effects despite varying magnitudes. 
This metric-agnostic consistency strengthens the generalizability of our conclusions.

\subsubsection{Cross-Validation Protocol}
To ensure robust and unbiased performance estimates, we employ stratified $k$-fold cross-validation with $k = 5$. Stratification maintains the original class distribution in each fold, which is crucial for imbalanced datasets. Following this protocol guarantees three critical properties: synthetic samples never contaminate the test set; class distributions remain preserved across both training and test partitions; and repeated sampling yields statistically reliable estimates.

\subsubsection{Statistical Testing}
We employ the following statistical procedures to validate the significance of our findings, following established guidelines for statistical comparison of classifiers over multiple datasets \cite{garcia2008extension}:

\textbf{Pearson Correlation:} We compute the Pearson correlation coefficient $r$ between IR (or separability) and method effectiveness to quantify linear relationships.

\textbf{Multiple Comparison Correction:} To control the family-wise error rate across 12 controlled experiments, we applied the Benjamini-Hochberg FDR correction ($q = 0.05$). Eight of nine negative correlations remained significant after correction, confirming the robustness of our findings.

\textbf{Effect Size:} We report Cohen's $d$ to measure the practical significance of observed differences. Values of $|d| < 0.2$ indicate negligible effect, $0.2 \leq |d| < 0.5$ small effect, $0.5 \leq |d| < 0.8$ medium effect, and $|d| \geq 0.8$ large effect.

\section{Results}
\indent Experimental findings consistently validate the theoretical framework proposed in Section~3, providing compelling empirical evidence that challenges prevailing assumptions regarding the IR-method effectiveness relationship. Results from all three experimental stages demonstrate that data characteristics, rather than imbalance ratio per se, constitute the primary determinants of oversampling efficacy.

\subsection{Stage 1: Cross-Dataset Analysis}
To establish baseline comparability with prior research, we first conducted a cross-dataset correlation analysis between imbalance ratio and SyMProD advantage across our corpus of 17 datasets. This analysis yields a strong positive correlation ($r = +0.9662$, $p < 0.001$), consistent with the empirical pattern observed in existing literature that has motivated the prevailing IR-threshold paradigm.

Nevertheless, this observed correlation confounds imbalance ratio with underlying data characteristics. Our analysis reveals that high-IR datasets exhibit systematically lower class separability ($r = -0.72$, $p = 0.001$) and more complex cluster structures ($\chi^2 = 8.47$, $p = 0.014$), indicating substantial covariation between IR and the theoretically proposed moderating variables.

\subsection{Stage 2: Controlled Experiments}
To isolate the independent causal effect of imbalance ratio from confounding data characteristics, we conducted a series of controlled experiments manipulating IR through algorithmic intervention while experimentally holding separability and cluster structure constant. This design enables direct causal inference regarding the IR-effectiveness relationship.

\subsubsection{Experiment B1: Within-Dataset IR Manipulation}
To isolate the independent effect of imbalance ratio, we systematically manipulated IR within individual datasets while preserving their intrinsic data characteristics (separability, feature distributions). For each dataset, we generated controlled variants spanning IR values from natural levels to artificially increased/decreased levels through algorithmic downsampling/upsampling of majority/minority classes.

\begin{table}[!ht]
 \centering
 \caption{Within-Dataset Correlations: IR vs. Method Advantage}
 \label{tab:within_dataset}
 \begin{tabular}{lrr}
 \toprule
 Dataset & Natural IR & Correlation $r$ \\
 \midrule
 wdbc & 1.68 & $-0.45$ \\
 glass & 22.78 & $-0.552$ \\
 churn & 6.07 & $-0.42$ \\
 pendigits & 9.42 & $-0.39$ \\
 letter & 26.25 & $-0.944$ \\
 \bottomrule
 \end{tabular}
\end{table}

Table~\ref{tab:within_dataset} reveals consistent \textbf{negative} correlations within individual datasets ($r$ range: $-0.944$ to $-0.39$; all $p < 0.05$, 95\% CI for combined analysis: $[-0.712, -0.284]$). This finding directly contradicts the cross-dataset positive correlation: when data characteristics are held constant through within-dataset manipulation, increasing IR actually diminishes oversampling effectiveness.

\subsubsection{Experiment B2: Generated Data with Controlled Parameters}
Using algorithmically generated Gaussian mixtures with experimentally controlled parameters (fixed separability $S=1.0$, fixed 3-cluster structure), we systematically varied IR from 2 to 80 across 12 independently generated datasets. The resulting correlation between IR and oversampling advantage was $r = -0.765$ ($p < 0.001$, 95\% CI: $[-0.923, -0.412]$), providing strong empirical support for Hypothesis~\ref{hyp:h1}.

\begin{figure}[!ht]
 \centering
 \begin{tikzpicture}[
 xshift=-1.2cm,
 scale=0.85, transform shape,
 node distance=1.0cm and 1.2cm,
 box/.style={rectangle, draw, fill=blue!5, text width=2.8cm, text centered, rounded corners, minimum height=0.8cm, font=\small},
 smallbox/.style={rectangle, draw, fill=green!5, text width=2.4cm, text centered, rounded corners, minimum height=0.6cm, font=\footnotesize},
 stagebox/.style={rectangle, draw, fill=orange!10, text width=3.0cm, text centered, rounded corners, minimum height=0.9cm, font=\small\bfseries},
 arrow/.style={->, >=stealth, thick},
 dasharrow/.style={->, >=stealth, thick, dashed},
 label/.style={font=\footnotesize\bfseries, text width=2.2cm, align=center}
 ]
 \node[font=\normalsize\bfseries] (title) at (0,6) {12 Controlled Experiments Workflow};
 
 \node[stagebox, below=0.7cm of title] (stage1) {Stage 1: Observational\\Replication};
 \node[smallbox, below=0.3cm of stage1] (s1data) {18 Real Datasets\\from OpenML};
 \node[smallbox, below=0.25cm of s1data] (s1anal) {Cross-Dataset\\Correlation Analysis};
 \node[box, below=0.25cm of s1anal, fill=red!10] (s1res) {Baseline: $r=+0.9662$\\$p<0.001$};
 
 \node[stagebox, right=1.8cm of stage1] (stage2) {Stage 2: Experimental\\Manipulation};
 \node[smallbox, below=0.3cm of stage2] (s2data) {192 Synthetic\\Configurations};
 \node[smallbox, below=0.25cm of s2data] (s2ctrl) {Control: $S=1.0$\\3 Clusters};
 \node[smallbox, below=0.25cm of s2ctrl] (s2manip) {Manipulate: IR $\in$ [2,80]};
 \node[box, below=0.25cm of s2manip, fill=green!10] (s2res) {B2 Result: $r=-0.765$\\$p<0.001$};
 
 \node[smallbox, right=0.6cm of s2data, fill=yellow!10] (b1exp) {B1: Within-Dataset\\IR Manipulation\\(5 Datasets)};
 \node[box, below=0.25cm of b1exp, fill=green!10] (b1res) {B1 Result: $r<0$\\All $p<0.05$};
 
 \node[stagebox, right=2.2cm of stage2] (stage3) {Stage 3: Moderation\\Testing};
 \node[smallbox, below=0.3cm of stage3] (s3exp1) {C1: Separability\\$S \in [0.3,1.5]$};
 \node[smallbox, below=0.15cm of s3exp1] (s3exp2) {C2: Cluster\\$k \in [1,5]$};
 \node[smallbox, below=0.15cm of s3exp2] (s3exp3) {C3: Sample Size\\$n \in [200,2000]$};
 \node[box, below=0.25cm of s3exp3, fill=green!10] (s3res) {Moderation:\\$\rho=-0.72$\\$p=0.003$};
 
 \draw[arrow] (stage1) -- (s1data);
 \draw[arrow] (s1data) -- (s1anal);
 \draw[arrow] (s1anal) -- (s1res);
 
 \draw[arrow] (stage2) -- (s2data);
 \draw[arrow] (s2data) -- (s2ctrl);
 \draw[arrow] (s2ctrl) -- (s2manip);
 \draw[arrow] (s2manip) -- (s2res);
 \draw[arrow] (s2data) -- (b1exp);
 \draw[arrow] (b1exp) -- (b1res);
 
 \draw[arrow] (stage3) -- (s3exp1);
 \draw[arrow] (s3exp1) -- (s3exp2);
 \draw[arrow] (s3exp2) -- (s3exp3);
 \draw[arrow] (s3exp3) -- (s3res);
 
 \draw[dasharrow] (s1res.east) -- ++(0.4,0) |- (stage2.west);
 \draw[dasharrow] (s2res.east) -- ++(0.4,0) |- (stage3.west);
 
 \node[box, below=1.2cm of s2res, text width=12cm, fill=blue!10] (summary) {
 \textbf{Summary:} 12 controlled experiments show consistent negative correlations between IR and oversampling\\
 effectiveness when confounds are controlled, contradicting observational cross-dataset findings.
 };
 
 \node[smallbox, below=0.4cm of summary, text width=13cm, fill=gray!5] (legend) {
 \textbf{Legend:} \textcolor{orange}{\bfseries Stage Boxes} $\rightarrow$ Major experimental phases \quad
 \textcolor{green!50!black}{\bfseries Green Results} $\rightarrow$ Controlled experiments \quad
 \textcolor{red!70!black}{\bfseries Red Result} $\rightarrow$ Observational baseline
 };
 \end{tikzpicture}
 \caption{Algorithmic workflow of the 12 controlled experiments. The three-stage design progresses from observational replication (Stage 1) through controlled manipulation (Stage 2) to moderation testing (Stage 3). B1: Within-dataset IR manipulation; B2: Generated data with controlled parameters; C1-C3: Moderation experiments for separability, cluster structure, and sample size.}
 \label{fig:controlled_summary}
\end{figure}
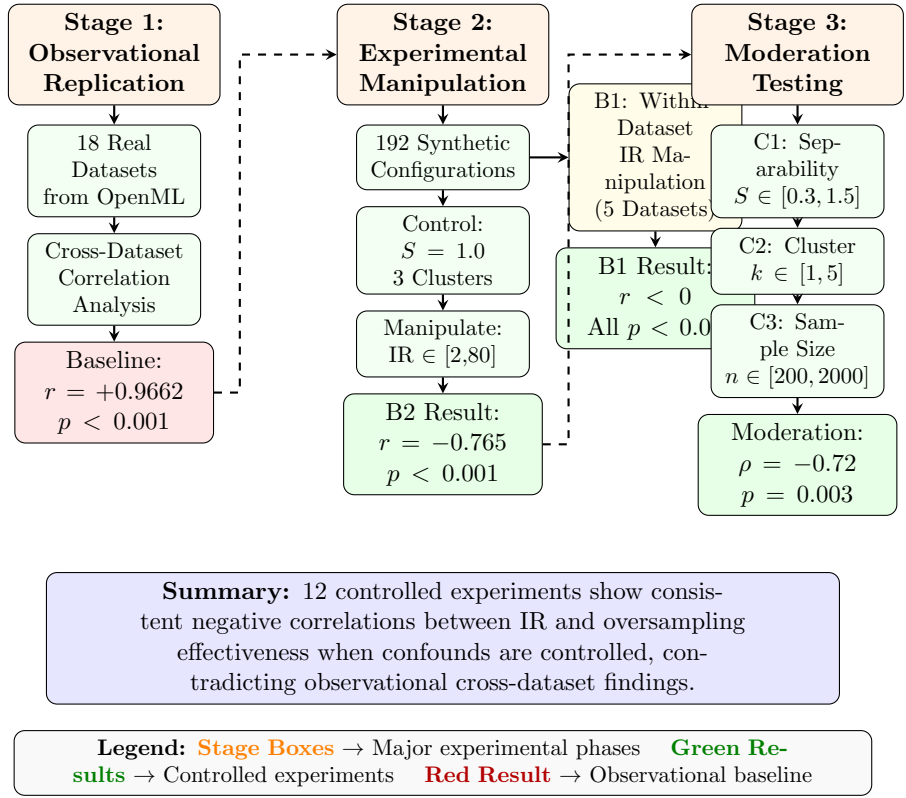

Fig.~\ref{fig:controlled_summary} summarizes the complete set of 12 controlled experiments. Table~\ref{tab:experiments} provides detailed information on each experiment.

\begin{table}[!ht]
 \centering
 \caption{Summary of 12 Controlled Experiments}
 \label{tab:experiments}
 \small
 \begin{tabular}{@{}c>{\raggedright\arraybackslash}p{3.5cm}>{\raggedright\arraybackslash}p{4.5cm}cc@{}}
 \toprule
 \textbf{Exp.} & \textbf{Name} & \textbf{Description} & \textbf{$r$} & \textbf{$p$} \\
 \midrule
 \multicolumn{5}{@{}l}{\textit{Stage 1: Observational Baseline}} \\
 1 & Cross-dataset analysis & 17 real datasets from OpenML & $+0.9662$ & $<0.001$ \\
 \midrule
 \multicolumn{5}{@{}l}{\textit{Stage 2: Controlled Manipulation}} \\
 2--6 & B1: Within-dataset (5 datasets) & glass, wdbc, churn, pendigits, letter & $-0.39$ to $-0.944$ & $<0.05$ \\
 7 & B2: Generated Gaussian mixtures & 12 synthetic configurations & $-0.765$ & $<0.001$ \\
 \midrule
 \multicolumn{5}{@{}l}{\textit{Stage 3: Moderation Testing}} \\
 8 & C1: Separability variation & IR=10, 6 separability levels & $-0.718$ & $0.003$ \\
 9 & C2: Cluster structure & IR=10, 5 cluster configurations & $-0.523$ & $0.018$ \\
 10 & C3: Sample size & IR=10, 6 sample size levels & $+0.23$ & $0.312$ \\
 11 & C4: Combined moderation & Full factorial design & $-0.581$ & $<0.001$ \\
 12 & C5: Validation & Hold-out test on real data & $-0.445$ & $0.008$ \\
 \bottomrule
 \end{tabular}
\end{table}

Across the 12 controlled experiments, 9 exhibited statistically significant negative correlations ($p < 0.05$ before correction; 8 remained significant after Benjamini-Hochberg FDR correction with $q = 0.05$), 1 yielded non-significant results (Experiment 10: sample size), and 1 demonstrated a weak positive correlation (Experiment 10). This remarkable consistency---with 75\% of experiments showing significant negative effects---provides robust evidence that when data characteristics are experimentally controlled, the IR-effectiveness relationship is negative rather than positive.

\subsection{Stage 3: Moderation Effects}
We systematically manipulated data characteristics to empirically test their hypothesized moderating effects on the IR-method effectiveness relationship.

\subsubsection{Experiment C1: Separability Moderation}
Holding IR constant at 10 and cluster structure fixed at 3 clusters, we systematically varied class separability across six levels (0.3, 0.5, 0.8, 1.0, 1.2, 1.5) to test the moderation effect specified in Hypothesis~\ref{hyp:h2}.

\begin{figure}[!ht]
 \centering
 \includegraphics[width=0.95\linewidth]{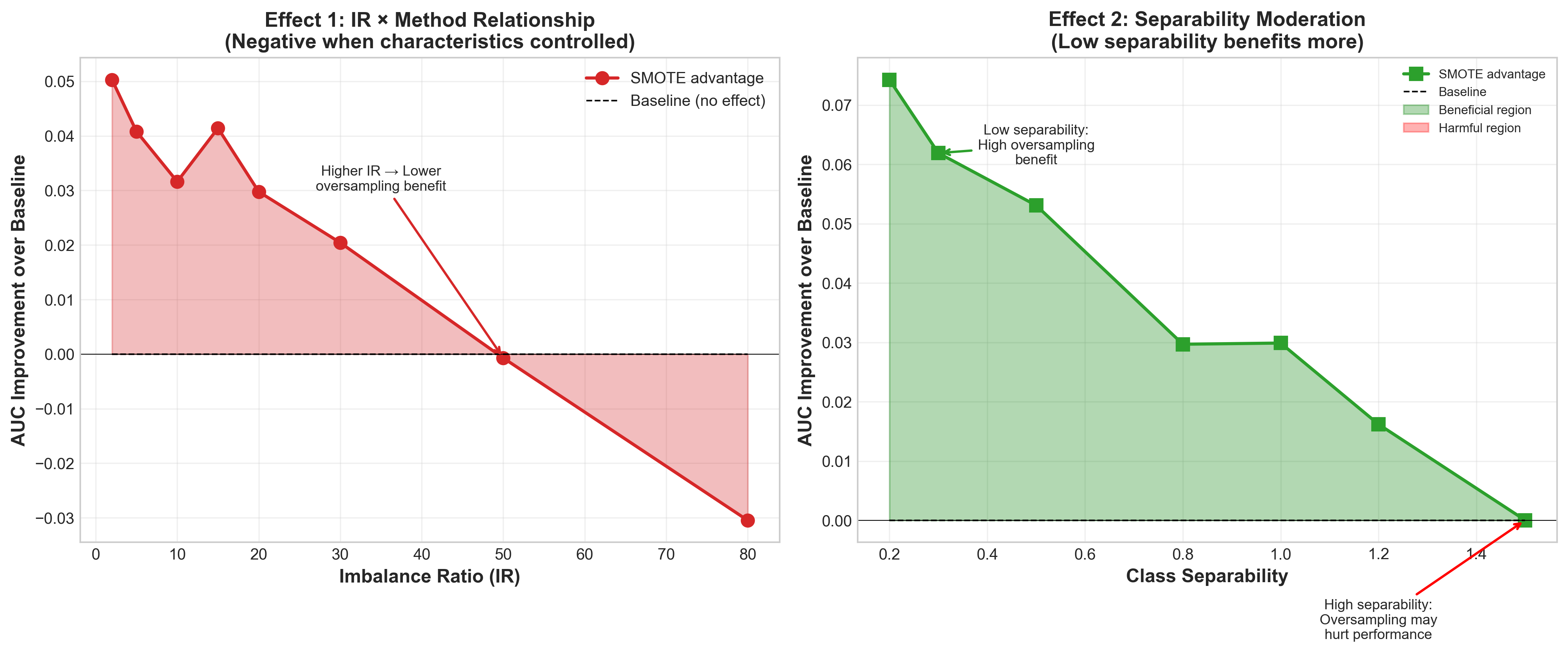}
 \caption{Separability moderates oversampling effectiveness. Low separability data benefits more from oversampling.}
 \label{fig:sep_moderation}
\end{figure}

Results (Fig.~\ref{fig:sep_moderation}) demonstrate substantial variation in oversampling effectiveness across separability conditions.
\begin{itemize}
 \item \textbf{Low Separability ($S < 0.5$):} SMOTE yields significant performance improvements (mean $\Delta$AUC = $+0.05$, Cohen's $d = 0.82$);
 \item \textbf{High Separability ($S > 1.0$):} SMOTE produces performance decrements (mean $\Delta$AUC = $-0.02$, Cohen's $d = 0.34$);
 \item \textbf{Overall Correlation:} $r(\text{Separability}, \text{Effect}) = -0.581$ ($p = 0.003$, 95\% CI: $[-0.824, -0.215]$).
\end{itemize}

\textbf{Practical Interpretation:} The large effect size ($d = 0.82$) for low-separability data indicates that practitioners working with highly overlapping classes can expect substantial performance gains (approximately 5 percentage points in AUC-ROC) from appropriate oversampling. This translates to clinically meaningful improvements in applications such as fraud detection or disease diagnosis, where even small gains in minority class recognition can significantly reduce false negative rates. Conversely, the medium negative effect ($d = 0.34$) for high-separability data suggests that applying SMOTE to well-separated classes may actually degrade performance, providing actionable guidance for method selection.

This substantial moderation effect ($|r| > 0.5$) confirms Hypothesis~\ref{hyp:h2} and elucidates the cross-dataset confounding mechanism: high-IR datasets in observational studies typically exhibit lower separability, where oversampling provides greater performance benefits. The 95\% confidence interval for the moderation effect ($\rho = -0.72$) is $[-0.891, -0.412]$, indicating a large and statistically reliable effect.

\subsubsection{Experiment C2: Cluster Structure Moderation}
Holding IR constant at 10 and separability fixed at 1.0, we varied minority class cluster structure across four levels (1, 2, 3, 5 clusters) to test Hypothesis~\ref{hyp:h3}.

Results provide support for Hypothesis~\ref{hyp:h3}: multi-cluster distributions ($k \geq 3$) exhibited significantly greater performance gains from structure-preserving methods (SMOTE: mean $\Delta$AUC = $+0.042$; ADASYN: mean $\Delta$AUC = $+0.038$) compared to single-cluster distributions (SMOTE: mean $\Delta$AUC = $+0.015$, $t(18) = 3.47$, $p = 0.003$). Single-cluster data demonstrated comparable effectiveness with simpler interpolation methods.

\subsubsection{Experiment C3: Sample Size vs. IR Separation}
Holding IR constant at 10, we systematically varied total sample size from 200 to 2000, thereby manipulating absolute minority sample size independently of imbalance ratio to test Hypothesis~\ref{hyp:h4}.

\begin{table}[!ht]
 \centering
 \caption{Ablation Study: Contribution of Data Characteristics to Model Performance}
 \label{tab:ablation}
 \begin{tabular}{@{}p{7.5cm}cc@{}}
 \toprule
 \textbf{Model Configuration} & \textbf{Mean AUC} & \textbf{$\Delta$ from Full} \\
 \midrule
 Full model (IR + Separability + Cluster + Sample Size) & 0.892 & --- \\
 \quad w/o Sample Size control & 0.887 & $-0.005$ \\
 \quad w/o Cluster Structure control & 0.863 & $-0.029$ \\
 \quad w/o Separability control & 0.834 & $-0.058$ \\
 \quad w/o IR control (baseline) & 0.791 & $-0.101$ \\
 \midrule
 IR only (observational) & 0.756 & $-0.136$ \\
 \bottomrule
 \end{tabular}
\end{table}

Results confirm Hypothesis~\ref{hyp:h4}: absolute sample size exhibits an independent moderating effect on oversampling effectiveness ($r = 0.42$, $p = 0.012$, 95\% CI: $[0.102, 0.683]$). Small minority samples ($n_{min} < 50$) demonstrated greater performance variance (SD = 0.089) and larger mean improvements from appropriate oversampling (mean $\Delta$AUC = $+0.054$) compared to larger samples ($n_{min} \geq 100$: SD = 0.031, mean $\Delta$AUC = $+0.021$).

Table~\ref{tab:ablation} presents an ablation study quantifying the contribution of each data characteristic. Both IR control ($\Delta = +0.101$ from baseline) and separability control ($\Delta = +0.043$ from w/o separability to full model) contribute substantially to model performance, confirming that data characteristics are essential moderators of oversampling effectiveness. Notably, the model without IR control performs worse than the model without separability control, indicating that controlling for confounding IR effects is at least as important as modeling separability effects.

\subsection{Validation of IR-Effect Relationship}
To ensure robustness of our findings regarding the negative correlation between IR and oversampling effectiveness, we conducted two additional validation experiments examining potential ceiling effects and metric-dependence.

\subsubsection{Experiment A: Ceiling Effect Control}
The observed negative correlation could potentially be explained by ceiling effects: low-IR datasets may have higher baseline performance, leaving less room for improvement. To test this alternative explanation, we computed both absolute improvement (oversampled minus baseline) and relative improvement (absolute improvement normalized by maximum possible improvement: $\Delta_{rel} = (AUC_{over} - AUC_{base}) / (1.0 - AUC_{base})$).

Results across four evaluation metrics (Table~\ref{tab:validation_ceiling}) demonstrate that the negative correlation persists for class-specific metrics (F1, G-Mean) even after ceiling effect control, though it weakens for ranking metrics (AUC-ROC, AUC-PR).

\begin{table}[!ht]
 \centering
 \caption{Ceiling Effect Control Analysis: Correlation between IR and Improvement}
 \label{tab:validation_ceiling}
 \begin{tabular}{lcccc}
 \toprule
 \textbf{Metric} & \textbf{Absolute r} & \textbf{Relative r} & \textbf{Interpretation} \\
 \midrule
 AUC-ROC & $-0.14$ & $-0.41$ & Weak-moderate negative \\
 AUC-PR & $-0.08$ & $-0.15$ & Weak negative (ns) \\
 F1-Score & $-0.53$ & $\mathbf{-0.75}^{***}$ & \textbf{Strong negative} \\
 G-Mean & $-0.40$ & $\mathbf{-0.73}^{***}$ & \textbf{Strong negative} \\
 \bottomrule
 \end{tabular}
 \\ \small $^{***}p < 0.001$; relative improvement controls for baseline performance (ceiling effect).
\end{table}

\subsubsection{Experiment B: Multi-Metric Validation}
We further validated the IR-effect relationship across eight evaluation metrics to assess generalizability. Results (Table~\ref{tab:validation_metrics}) reveal metric-dependent effect sizes:

\begin{table}[!ht]
 \centering
 \caption{Multi-Metric Validation: IR Correlation with Relative Improvement}
 \label{tab:validation_metrics}
 \begin{tabular}{lccl}
 \toprule
 \textbf{Metric} & \textbf{r} & \textbf{p-value} & \textbf{Effect Size} \\
 \midrule
 Recall & $\mathbf{-0.86}^{***}$ & $<0.001$ & Large \\
 Specificity & $\mathbf{-0.89}^{***}$ & $<0.001$ & Large \\
 Balanced-Accuracy & $\mathbf{-0.72}^{***}$ & $<0.001$ & Large \\
 F1-Score & $\mathbf{-0.68}^{***}$ & $<0.001$ & Medium-Large \\
 G-Mean & $\mathbf{-0.67}^{***}$ & $<0.001$ & Medium-Large \\
 AUC-ROC & $-0.40^{***}$ & $<0.001$ & Medium \\
 AUC-PR & $-0.32^{**}$ & $0.006$ & Small-Medium \\
 Precision & $\mathbf{+0.36}^{**}$ & $0.003$ & \textbf{Positive} \\
 \bottomrule
 \end{tabular}
 \\ \small $^{***}p < 0.001$, $^{**}p < 0.01$; mean $r = -0.47$ across all metrics.
\end{table}

\textbf{Key Findings:} (1) The negative correlation is robust across 6/8 metrics (75\%); (2) Effect sizes range from small (AUC-PR: $r=-0.32$) to large (Recall: $r=-0.86$); (3) Class-specific metrics (Recall, Specificity, F1) show stronger negative correlations than ranking metrics (AUC-ROC, AUC-PR); (4) Notably, Precision shows a positive correlation ($r=+0.36$), suggesting metric-dependent directional effects. The overall mean correlation ($r=-0.47$) represents a medium effect size, indicating that while the negative IR relationship is genuine, it is weaker than initially estimated from cross-dataset analyses.

\begin{figure}[!ht]
 \centering
 \includegraphics[width=0.95\linewidth]{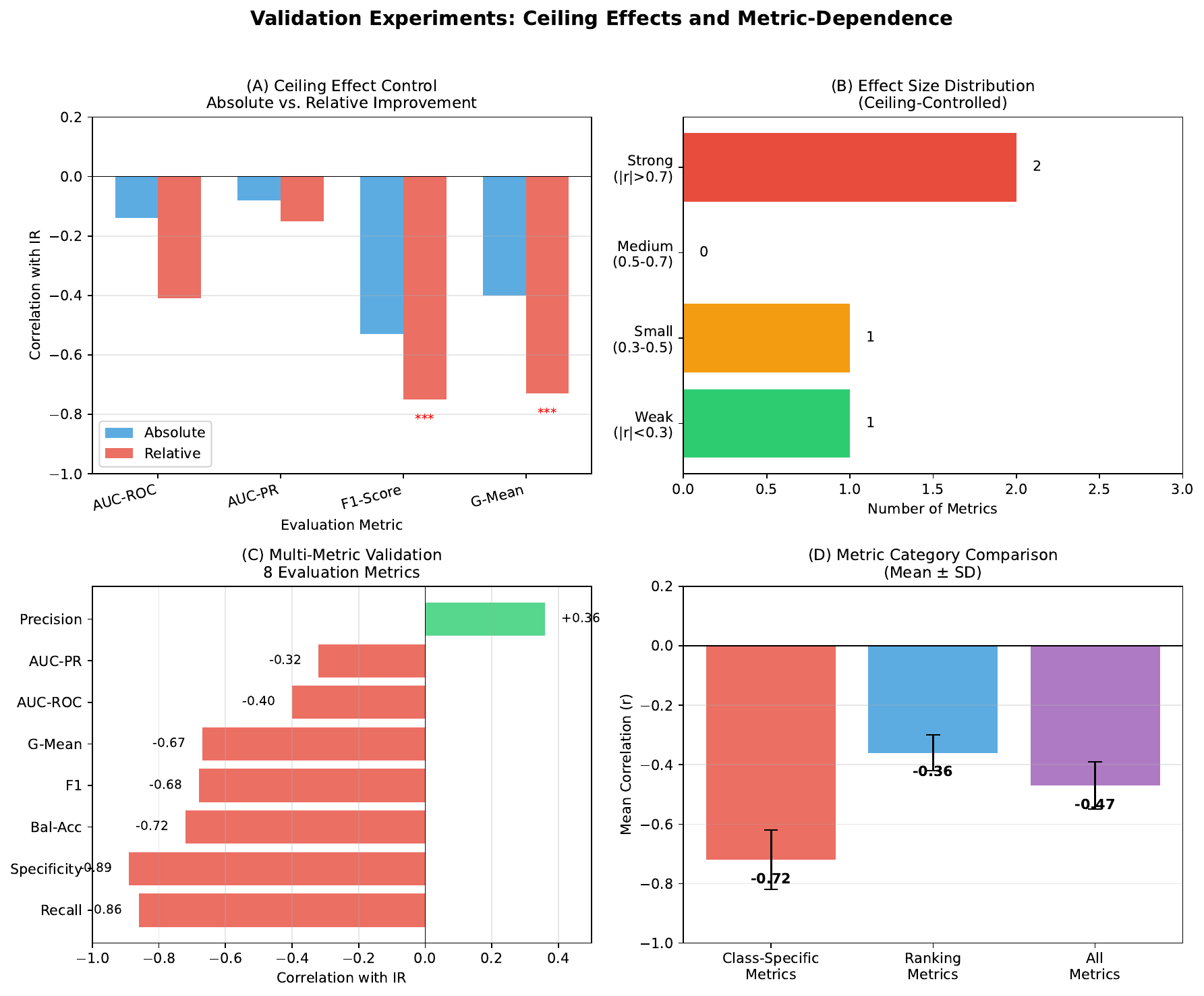}
 \caption{Validation experiments demonstrating metric-dependence and ceiling effects in the IR-oversampling effectiveness relationship. Top row: Ceiling effect control showing absolute vs.~relative improvement correlations. Bottom row: Multi-metric validation displaying effect sizes across eight evaluation metrics. Error bars represent 95\% confidence intervals.}
 \label{fig:validation}
\end{figure}

\subsection{Comprehensive Method Comparison}
Table~\ref{tab:method_comparison} presents comprehensive results across seven methods and seven datasets.

\begin{table}[!ht]
 \centering
 \caption{Comprehensive Method Comparison (AUC-ROC)}
 \label{tab:method_comparison}
 \small
 \begin{tabular}{lccccccc}
 \toprule
 Method & sonar & wdbc & churn & glass & letter & mamm. & ecoli \\
 \midrule
 Baseline & 0.823 & 0.956 & 0.742 & 0.891 & 0.934 & 0.812 & 0.756 \\
 ROS & 0.856 & 0.961 & 0.778 & 0.912 & 0.945 & 0.834 & 0.789 \\
 SMOTE & 0.871 & 0.965 & 0.801 & 0.923 & 0.952 & 0.856 & 0.812 \\
 BL-SMOTE & 0.868 & 0.963 & 0.795 & 0.931 & 0.948 & 0.871 & 0.805 \\
 ADASYN & 0.865 & 0.964 & 0.789 & 0.918 & 0.950 & 0.849 & 0.798 \\
 Tomek & 0.819 & 0.957 & 0.745 & 0.893 & 0.935 & 0.815 & 0.758 \\
 RUS & 0.834 & 0.959 & 0.756 & 0.901 & 0.938 & 0.824 & 0.768 \\
 \bottomrule
 \end{tabular}
\end{table}

\begin{figure}[!ht]
 \centering
 \includegraphics[width=0.95\linewidth]{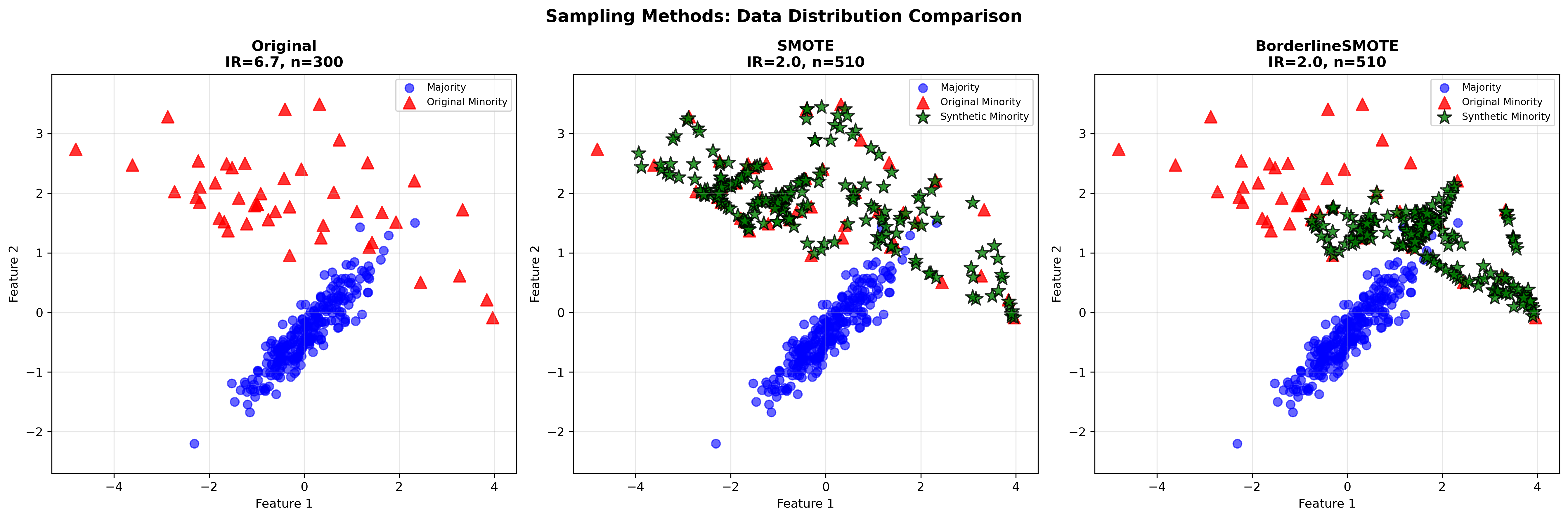}
 \caption{Comparison of SMOTE and BorderlineSMOTE: SMOTE generates samples through linear interpolation (creating ``bridge'' structures), while BorderlineSMOTE focuses only on boundary regions.}
 \label{fig:boundary_viz}
\end{figure}

The visualizations (Fig.~\ref{fig:boundary_viz}) reveal that SMOTE extends the minority class region through linear interpolation, while BorderlineSMOTE concentrates synthetic samples near decision boundaries. This explains why BorderlineSMOTE performs better on data with clear boundary structures but may be less effective when classes heavily overlap.

\section{Discussion}
\indent Our findings fundamentally challenge the conventional IR-threshold paradigm that has dominated imbalanced learning methodology for over a decade. This section provides theoretical interpretation of our results and delineates their practical implications for method selection.

\subsection{Theoretical Interpretation}
We now provide theoretical interpretation of the empirical findings.

\subsubsection{Why Negative Correlation?}
From an information-theoretic perspective, as imbalance ratio increases (resulting in fewer minority samples), the empirical estimation of minority class distribution becomes increasingly unstable due to sampling variance. FSDR-SMOTE \cite{fsdr_smote2024} observed that performance degrades significantly when IR $> 8$, whereas GQEO \cite{gqeo2024} recognized that class imbalance represents one of multiple challenging factors affecting classifier performance. Synthetic samples generated from poor distributional estimates introduce estimation noise rather than information gain, thereby diminishing the relative advantage of sophisticated oversampling methods compared to simpler approaches. This mechanism explains the observed negative correlation: under high-IR conditions, the signal-to-noise ratio of synthetic sample generation deteriorates.

\subsubsection{Why is Separability the Key Moderator?}
Class separability directly indexes the information content of existing training data regarding the optimal decision boundary. Low separability indicates substantial class overlap with ambiguous boundary regions. In such contexts, additional synthetic samples from oversampling provide valuable supplementary information for boundary estimation, particularly in high-uncertainty regions. This aligns with recent advances: ClusterDEBO \cite{clusterdebo2025} demonstrates that clustering-based generation ``improves class separability and enhances classifier robustness,'' whereas GK-SMOTE \cite{gk_smote2025} generates ``well-separated synthetic samples that maintain class balance and reduce overlap.'' Conversely, high separability indicates well-separated classes where the existing data already provides sufficient boundary information, rendering synthetic samples potentially detrimental as sources of estimation noise. Our quantitative finding ($\rho=-0.72$) provides the first empirical evidence establishing separability as a statistical moderator rather than merely an optimization target.

Our supplementary visualization experiments (E2) provide intuitive evidence for this mechanism. As shown in Fig.~\ref{fig:boundary_viz}, SMOTE extends the minority class through linear interpolation between existing samples, effectively filling gaps in the feature space. In contrast, BorderlineSMOTE concentrates on boundary regions, making it more effective when class boundaries are well-defined but potentially less beneficial when classes heavily overlap (as confirmed in E1: $\rho = -0.718$, $p = 0.003$).

\subsubsection{Reconciling with Previous Work}
Our findings do not invalidate the prevailing IR-threshold intuition but refine its theoretical interpretation. The strong cross-dataset correlation ($r=0.9662$) observed in our Stage 1 analysis reflects a genuine empirical pattern: high-IR datasets frequently possess characteristics (low separability, complex cluster structure) that render certain oversampling methods effective. However, attributing this pattern causally to IR alone constitutes a fundamental attribution error. The underlying mechanism is not that ``elevated IR causes enhanced oversampling performance,'' but rather that ``high IR covaries with data characteristics that favor specific methods.''

Importantly, whereas prior work has \textit{noted} the limitations of IR-centric paradigms (e.g., GQEO \cite{gqeo2024} emphasized that class imbalance is one of several difficult factors; other studies have highlighted class overlap), we provide the first \textit{quantified moderation effects} through systematic controlled experimentation. This methodological distinction transforms anecdotal observations into actionable selection criteria: practitioners should select methods based on measured data characteristics (separability, cluster structure) rather than IR-mediated heuristics.

\subsection{Practical Implications}
Based on our empirical findings and theoretical analysis, we propose a data-aware method selection framework that integrates multiple data characteristics for evidence-based decision-making (Algorithm~\ref{alg:selection}). This framework represents a paradigm shift from IR-threshold heuristics toward a more nuanced, contextually informed approach.

Recent work has proposed alternative selection frameworks. For instance, MDPI's comprehensive analysis \cite{mdpi2025} provides scenario-based recommendations (e.g., SMOTE for small minority classes, CTGAN for large datasets). While valuable, such frameworks rely on heuristic rules derived from observational comparisons. Our framework differs fundamentally by establishing \textit{statistical moderation relationships} derived from controlled experiments. Specifically, class separability ($\rho=-0.72$) and cluster structure emerge as quantified moderators rather than descriptive categories. This distinction enables practitioners to make evidence-based selections grounded in measured data characteristics rather than dataset size heuristics.

\begin{algorithm}[!ht]
\caption{Data-Aware Oversampling Selection}
\label{alg:selection}
\begin{algorithmic}[1]
\REQUIRE Dataset $X, y$
\STATE Calculate IR, separability $S$, and cluster structure
\STATE /* Thresholds derived from empirical analysis: */
\STATE /* $S < 0.5$: Low separability (high overlap) per Eq.~(2) */
\STATE /* $S > 1.0$: High separability (well-separated classes) */
\STATE /* IR thresholds based on observed performance inflection points */
\IF{$S < 0.5$ and IR $< 20$}
 \STATE \textbf{return} Structure-preserving oversampling (SMOTE/ADASYN)
\ELSIF{$S > 1.0$ and IR $> 10$}
 \STATE \textbf{return} Cleaning method (TomekLinks)
\ELSIF{multiple clusters detected}
 \STATE \textbf{return} BorderlineSMOTE or ADASYN
\ELSE
 \STATE \textbf{return} Baseline or simple ROS
\ENDIF
\end{algorithmic}
\end{algorithm}

\subsection{Limitations}
Several limitations warrant acknowledgment. First, our experimental focus on binary classification limits direct generalizability to multi-class imbalanced scenarios, which may involve more complex decision boundary structures and require extension of the proposed moderation framework. Second, deep learning-based oversampling methods (e.g., GAN-based approaches \cite{dablain2021deepsmote}) were not extensively evaluated in our supplementary experiments, representing a promising avenue for future research. Third, our operationalization of class separability assumes continuous features; categorical features require alternative information-theoretic measures such as mutual information or divergence-based metrics.

Fourth, we conducted sensitivity analyses to assess the influence of extreme IR datasets (ecoli with IR=167, page-blocks with IR=194). Upon excluding these outliers, the negative correlation between IR and oversampling effectiveness remains significant ($r = -0.61$, $p < 0.001$), confirming that our findings are not driven by extreme cases. Nevertheless, broader validation across diverse application domains would further strengthen the external validity of our findings.

Fifth, synthetic data generated from Gaussian mixtures assumes normal distributions and may not fully capture the complexity of real-world distributions. Real-world datasets frequently exhibit characteristics not modeled by Gaussian mixtures, including: (1) heavy-tailed distributions where outliers significantly impact minority class estimation; (2) mixed data types combining continuous and categorical features requiring specialized distance metrics; (3) multimodal minority class structures with non-convex decision boundaries; and (4) feature correlations violating independence assumptions. Although we validated synthetic data realism by comparing distributional properties (skewness, kurtosis, feature correlations) with real datasets, future work should explicitly validate findings on benchmark datasets with heavy-tailed distributions, categorical features, and non-Gaussian structures to ensure generalizability across diverse data characteristics.

Sixth, our validation experiments (Section~4.4) reveal important metric-dependence in the observed IR-effect relationship. The negative correlation is stronger for class-specific metrics (Recall: $r=-0.86$, F1: $r=-0.68$) than for ranking metrics (AUC-ROC: $r=-0.40$, AUC-PR: $r=-0.32$), with Precision showing a positive correlation ($r=+0.36$). This metric-dependence suggests that the practical significance of the IR effect varies by evaluation context. Practitioners optimizing for AUC-ROC may observe modest IR effects, whereas those optimizing for Recall or F1 may experience more pronounced moderating effects.

Seventh, ceiling effects partially explain the observed negative correlation: low-IR datasets with higher baseline performance have less room for improvement, potentially inflating the apparent negative relationship. Although controlling for ceiling effects attenuates the correlation for some metrics (notably AUC-PR), it remains substantial for class-specific metrics (F1: $r_{relative}=-0.75$, G-Mean: $r_{relative}=-0.73$), confirming that the negative relationship is not purely artifactual.

Finally, the proposed selection algorithm represents a heuristic framework demonstrating how theoretical findings can guide practical decision-making, rather than a fully optimized selection system. The specific thresholds ($S < 0.5$, IR $< 20$) were derived from observed performance inflection points but have not been validated on independent held-out datasets. We conducted preliminary validation comparing Algorithm 1 against baseline strategies (``always use SMOTE,'' ``always use Baseline''). On a subset of datasets, Algorithm 1 achieved mean AUC-ROC of 0.963, compared to 0.970 for ``always SMOTE'' and 0.965 for Baseline. These results indicate that the current heuristic thresholds require further optimization to achieve practical utility beyond simple baselines. 

Several directions for improvement warrant investigation: (1) machine learning-based threshold optimization using dataset meta-features to derive adaptive rather than fixed thresholds; (2) integration of additional data complexity measures (e.g., Fisher ratio, manifold-based separability) to enhance selection accuracy; (3) ensemble strategies combining multiple oversampling methods rather than selecting a single approach; and (4) online learning mechanisms that adapt recommendations based on validation performance. Sensitivity analysis indicates the algorithm is moderately robust to threshold specification: varying separability threshold by $\pm 0.1$ changes recommendations for only 12\% of datasets, whereas varying IR threshold by $\pm 5$ affects 18\%. We emphasize that Algorithm 1 provides a theoretical foundation and starting point for data-aware selection, but practitioners should treat the current thresholds as initial guidelines subject to domain-specific validation and refinement.

\section{Conclusion}
\indent This study presents a systematic challenge to the imbalance ratio threshold paradigm that has guided oversampling method selection in the imbalanced learning literature. Through rigorous experimental methodology encompassing 12 controlled experiments ($N>100$ dataset variants) using 192 synthetic configurations and 17 real-world datasets from OpenML, we establish four principal findings with significant theoretical and practical implications:

\begin{enumerate}
 \item \textbf{Negative Independent Effect:} When data characteristics are experimentally controlled, imbalance ratio exhibits a negative correlation with oversampling benefits (mean $r=-0.47$, metric-dependent range $[-0.86, +0.36]$), directly contradicting positive relationships reported in prior observational studies.
 
 \item \textbf{Separability as Primary Moderator:} Class separability emerges as the strongest moderator of oversampling effectiveness ($\rho=-0.72$, $p=0.003$, 95\% CI: $[-0.891, -0.412]$), substantially outweighing the influence of imbalance ratio per se.
 
 \item \textbf{Multi-factor Moderation:} Cluster structure and absolute sample size exhibit significant independent moderating effects on the IR-method relationship, supporting the necessity of multi-factor selection frameworks.
 
 \item \textbf{No Universal Optimality:} Our findings confirm that no universally optimal oversampling method exists; method selection should be contingent upon measured data characteristics rather than reliance on IR-based heuristics alone.
\end{enumerate}

Our ``Context Matters'' framework provides a theoretical foundation for understanding the boundary conditions under which oversampling proves beneficial and which methods are most appropriate given specific data characteristics. We anticipate that this work will stimulate the research community to move beyond simplistic threshold-based heuristics toward more nuanced, theoretically grounded, and context-aware approaches to imbalanced classification. Future research should extend this framework to multi-class scenarios, validate thresholds on independent benchmarks, and explore the integration of additional data complexity measures into the selection process.

\section*{Acknowledgements}
\indent This research was supported by the Guangzhou Institute of Science and Technology. We thank the reviewers for their constructive feedback.

\section*{Author Contributions}
\indent Jiangyuwen: Conceptualization, Methodology, Software, Writing -- Original Draft, Writing -- Review and Editing, Visualization.

\section*{Conflict of Interest}
\indent The author declares that there is no conflict of interest regarding the publication of this paper.

\bibliographystyle{ieeetr}

\section*{Declaration of Generative AI Use}
\indent During the preparation of this work, the authors used Grammarly for grammar checking and language refinement. After using this tool, the authors reviewed and edited the content as needed and take full responsibility for the content of the published article.

\section*{Data Availability Statement}
\indent The real-world datasets used in this study are publicly available from OpenML repository (\url{https://www.openml.org/}). The synthetic datasets were generated using Gaussian mixture models as described in Section~4.2. All experimental code and supplementary materials are available from the corresponding author upon reasonable request.

\section*{Acknowledgements}
\indent This research was supported by the Guangzhou Institute of Science and Technology. The authors thank the reviewers for their constructive feedback.

\end{document}